\documentclass{ceurart}

\usepackage{booktabs}
\usepackage{enumitem}
\usepackage{graphicx}

\begin{document}

\copyrightyear{2026}
\copyrightclause{Copyright for this paper by its authors. Use permitted under Creative Commons License Attribution 4.0 International (CC BY 4.0).}
\conference{ISWC 2026 Companion Volume, October 25--29, 2026, Bari, Italy}

\title{OntoInk: Interactive Ontology Visualization, Validation, and Reasoning}

\author[1,2]{Ebrahim Norouzi}[
  orcid=0000-0002-2691-6995,
  email=ebrahim.norouzi@fiz-karlsruhe.de,
]
\cormark[1]

\author[1]{Jörg Waitelonis}[
  orcid=0000-0001-7192-7143,
  email=joerg.waitelonis@fiz-karlsruhe.de,
]

\author[1,2]{Harald Sack}[
  orcid=0000-0001-7069-9804,
  email=harald.sack@fiz-karlsruhe.de,
]

\address[1]{FIZ Karlsruhe -- Leibniz Institute for Information Infrastructure, Germany}
\address[2]{Karlsruhe Institute of Technology (KIT), Institute AIFB, Germany}

\cortext[1]{Corresponding author.}

\begin{abstract}
Ontology documentation, visualization, and validation are usually carried out with separate tools. This split workflow slows down development and makes knowledge transfer harder. We present OntoInk, an open-source MkDocs plugin that brings these activities together. Within a single documentation-as-code pipeline, OntoInk renders interactive ontology diagrams, validates instance data against SHACL shapes, runs OWL\,DL reasoning, and supports inline Turtle editing. General-purpose diagram plugins for MkDocs cannot parse RDF, dereference IRIs, overlay SHACL constraints, or run OWL reasoning. Compared with standalone ontology visualization tools, OntoInk embeds interactive and editable diagrams directly into documentation pages. A live demo and source code are available at \url{https://ise-fizkarlsruhe.github.io/ontoink/}.

\end{abstract}


\maketitle

\section{Introduction}\label{sec:intro}

Ontologies form the formal schema layer of knowledge graphs, yet their documentation and visualization are usually carried out with different tools. A wide range of ontology visualization tools has appeared over the last decade~\cite{dudas2018survey,gomes2024ontosurvey,lohmann2015webvowl,falco2014graffoo,chavez2022chowlk}, but they produce either static diagrams or stand-alone web applications that sit apart from the documentation workflow---conceptual models drawn in one tool must be kept in sync with the OWL source by hand. HTML documentation generators such as Widoco~\cite{garijo2017widoco} embed a (static) WebVOWL diagram, and vocabulary-development environments such as VoCol~\cite{halilaj2016vocol} add version control and documentation generation to CI pipelines---but their output still lacks interactive graph exploration, SHACL constraint visualization, and OWL reasoning.

In parallel, the documentation-as-code idea has been widely adopted, and static site generators such as MkDocs are commonly extended with diagram plugins (D2~\cite{d2}, Mermaid.js~\cite{mermaid}). These general-purpose approaches, however, are \emph{not ontology-aware}: they cannot parse RDF, resolve IRIs to fetch labels, overlay SHACL constraints, validate instance data, or execute OWL reasoning.

OntoInk\footnote{OntoInk source code: \url{https://github.com/ISE-FIZKarlsruhe/ontoink}} closes this gap. It treats the ontology source as the \emph{single point of truth} from which visualization, validation, and reasoning are derived in a Markdown-based pipeline. RDF Turtle~\cite{beckett2014turtle} files are embedded in Markdown through fenced code blocks. At build time, the RDF graph is parsed, SHACL~\cite{knublauch2017shacl} constraints are extracted, and OWL\,DL reasoning is executed. At runtime, the page shows an interactive diagram with SHACL constraint overlays, a live RDF Turtle editor with syntax highlighting, and publication-ready export.

The contributions of this work are: (1)~a documentation-as-code architecture that integrates ontology visualization, SHACL validation, OWL reasoning, and inline editing in a single MkDocs pipeline; (2)~a formal visual notation that overlays SHACL constraints with cardinality badges directly on the graph; and (3)~a visual SHACL shape editor that authors shapes through a form interface.

\section{Related Work}\label{sec:related}

Table~\ref{tab:comparison} positions OntoInk against representative tools across three categories. \emph{Ontology visualization tools} focus on rendering ontologies as diagrams or interactive visualizations. Representative examples include WebVOWL~\cite{lohmann2015webvowl}, which implements the VOWL notation~\cite{lohmann2014vowl}, Graffoo~\cite{falco2014graffoo}, OWL2Graffoo~\cite{duchateau2026owl2graffoo}, Chowlk~\cite{chavez2022chowlk}, OWLGrEd~\cite{barzdins2010owlgred}, and the OntoDia~\cite{mouromtsev2015ontolia}/Reactodia~\cite{reactodia} family. These tools typically render ontologies as static diagrams or stand-alone web applications that are separate from the documentation workflow. None combine SHACL overlays, OWL reasoning, and inline editing. Among them, only UnSHACLed~\cite{lieber2022unshacled} supports SHACL, but its focus is on constraint authoring rather than documentation.

\emph{Documentation generators} such as Widoco~\cite{garijo2017widoco}, LODE~\cite{peroni2012lode}, and pyLODE~\cite{car2022pylode} generate read-only HTML documentation from ontologies. \emph{Documentation-as-code} platforms, including VoCol~\cite{halilaj2016vocol} and OnToology~\cite{alobaid2019ontoology}, extend this approach by integrating version control and continuous integration. However, they still produce static documentation pages rather than interactive documentation environments. General-purpose diagram plugins for MkDocs (D2, Mermaid) are \emph{not ontology-aware}: they cannot parse RDF, dereference IRIs, overlay SHACL constraints, or run reasoning, so their diagrams drift from the source like hand-drawn figures.

As shown in Table~\ref{tab:comparison}, OntoInk, to the best of our knowledge, is the first tool that combines interactive ontology visualization with formal SHACL constraint overlays, OWL reasoning, IRI dereferencing, and inline Turtle editing within a documentation-as-code workflow.

\begin{table}[t]
\caption{Feature comparison of OntoInk with related ontology visualization and documentation tools. $\bullet$ = supported, $\circ$ = partial, -- = not supported.}\label{tab:comparison}
\centering
\small
\begin{tabular}{@{}lcccccc@{}}
\toprule
 & \rotatebox{70}{Interactive Vis.}
 & \rotatebox{70}{SHACL Overlay}
 & \rotatebox{70}{OWL Reasoning}
 & \rotatebox{70}{Inline Editing}
 & \rotatebox{70}{Doc. Integration}
 & \rotatebox{70}{Pub.-Ready Export} \\
\midrule
Prot\'{e}g\'{e}~\cite{musen2015protege}         & $\circ$   & $\circ$   & $\bullet$ & $\bullet$ & --        & -- \\
WebVOWL~\cite{lohmann2015webvowl}                & $\bullet$ & --        & --        & --        & --        & $\circ$ \\
WebVOWL Editor~\cite{wiens2018webvowl}           & $\bullet$ & --        & --        & $\bullet$ & --        & $\circ$ \\
Graffoo~\cite{falco2014graffoo}                  & --        & --        & --        & --        & --        & $\bullet$ \\
OWL2Graffoo~\cite{duchateau2026owl2graffoo}      & $\circ$   & --        & --        & $\bullet$ & $\circ$   & $\bullet$ \\
Chowlk~\cite{chavez2022chowlk}                  & --        & --        & --        & $\circ$   & --        & $\bullet$ \\
OWLGrEd~\cite{barzdins2010owlgred}              & $\circ$   & --        & --        & $\bullet$ & --        & $\circ$ \\
OntoDia~\cite{mouromtsev2015ontolia}             & $\bullet$ & --        & --        & --        & --        & -- \\
Reactodia~\cite{reactodia}                       & $\bullet$ & --        & --        & $\bullet$ & --        & $\circ$ \\
UnSHACLed~\cite{lieber2022unshacled}             & $\circ$   & $\bullet$ & --        & $\bullet$ & --        & -- \\
Widoco~\cite{garijo2017widoco}                   & $\bullet$ & --        & --        & --        & $\bullet$ & -- \\
pyLODE~\cite{car2022pylode}                      & --        & --        & --        & --        & $\bullet$ & -- \\
VoCol~\cite{halilaj2016vocol}                    & --        & --        & --        & --        & $\bullet$ & -- \\
OnToology~\cite{alobaid2019ontoology}            & $\circ$   & --        & --        & --        & $\bullet$ & -- \\
\textbf{OntoInk}                                 & $\bullet$ & $\bullet$ & $\bullet$ & $\bullet$ & $\bullet$ & $\bullet$ \\
\bottomrule
\end{tabular}
\end{table}

\section{System Overview}\label{sec:system}

OntoInk is implemented as an MkDocs plugin with a two-phase architecture: a \emph{build-time} processing pipeline in Python and a \emph{runtime} interactive layer in JavaScript.

\subsection{Build-Time Processing}\label{sec:buildtime}

The plugin is triggered when MkDocs encounters an \texttt{ontoInk} fenced code block. The code block specifies a Turtle data file, an optional SHACL shapes file, and visualization parameters. The plugin then executes the following processing pipeline:

\begin{enumerate}
\item \textbf{RDF Parsing \& Classification.} The Turtle source is loaded via \texttt{rdflib}~\cite{krech2023rdflib}. Nodes are classified as classes, individuals, literals, or datatypes from \texttt{rdf:type}/\texttt{rdfs:subClassOf} axioms, each receiving a distinct visual notation.

\item \textbf{SHACL Extraction \& Validation.} \texttt{sh:NodeShape} instances are traversed; property constraints (\texttt{sh:path}, \texttt{sh:minCount}, \texttt{sh:maxCount}, \texttt{sh:class}) are mapped to graph edges with cardinality badges. The data graph is then validated against the shapes graph using pySHACL~\cite{car2020pyshacl}, extracting violations with focus node, property path, severity, and message.

\item \textbf{OWL Reasoning.} Four reasoner backends can be selected per page or request: HermiT~\cite{glimm2014hermit} via \texttt{owlready2}~\cite{lamy2017owlready} (the default for full OWL\,DL inference if Java is available); Konclude~\cite{steigmiller2014konclude}, an optimized native tableau reasoner, as a Java-free alternative; the same Konclude kernel compiled to WebAssembly via \texttt{rdf-reasoner-konclude}~\cite{hanke2026rdfreasoner}, which lets OntoInk run OWL\,DL reasoning entirely in the browser; and, as a lightweight pure-Python fallback, the OWL\,RL profile~\cite{owl2profiles}.
\end{enumerate}

\subsection{Runtime Interaction}\label{sec:runtime}

At page load, the embedded data is decoded, and an interactive graph is initialized. The runtime layer provides these capabilities.

\paragraph{Visualization, Navigation and Export.}
Ontology graphs are rendered using a formal visual notation inspired by established approaches~\cite{lohmann2014vowl,dudas2018survey}. Classes are \emph{automatically color-coded by ontology source namespace}, enabling immediate identification of cross-ontology dependencies. An \emph{Edit Layout} panel lets users recolor nodes per \emph{declared namespace}, change node shapes, and restyle edges, with the customizations carried into the exported image. Node clicks open popups with IRI dereferencing via content negotiation, extracting labels, comments, types, and deprecation status from resolved resources. Inferred triples are shown in a dedicated panel and can be overlaid on the graph as purple dotted edges, so readers can tell asserted statements apart from derived ones. A \emph{Validate with Inferences} action re-runs SHACL validation over the union of asserted and inferred triples, revealing constraints satisfied only after reasoning. Diagram export is available in publication-ready formats for direct inclusion in publications.

\paragraph{SHACL Overlay, Inline Editing and Authoring.}
SHACL constraints are overlaid as bold dashed edges with cardinality badges $[\mathit{min}..\mathit{max}]$, letting engineers inspect structural requirements directly on the graph without reading raw SHACL. An integrated Turtle editor with syntax highlighting enables editing instance data in place; on update, the Turtle is re-parsed, the graph rebuilt, and the data re-validated \emph{entirely in the browser}. This client-side engine covers the full SHACL Core specification---value-type, cardinality, value-range, string, property-pair, logical, and shape-based constraints. Beyond inspecting existing shapes, OntoInk provides a dedicated SHACL shape editor that lowers the barrier to writing constraints. Users create shapes through a form-based interface instead of editing Turtle directly. They first select a target class and then add property constraints, such as the property path, cardinality, datatype, node kind, or pattern. The corresponding SHACL Turtle representation is generated automatically.

\section{Deployment and Use Cases}\label{sec:usecases}

OntoInk has been deployed within the German National Research Data Infrastructure (NFDI)~\cite{sure2019nfdi}, a large-scale initiative for cross-disciplinary research data management.

\paragraph{NFDI-MatWerk}
The MatWerk Knowledge Graph (MatWerk KG)\footnote{The MatWerk KG: \url{https://nfdi.fiz-karlsruhe.de/matwerk/}, the SHACL shapes are documented using OntoInk and are available at \url{https://nfdi.fiz-karlsruhe.de/matwerk/shapes/}}~\cite{msekg_zenodo_v211} is built on the NFDI MatWerk Ontology (MWO)~\cite{mwo2026}, a BFO-compliant~\cite{arp2015bfo} ontology for research data management in materials science and engineering. On top of MWO, the KG introduces SHACL shapes that enforce data quality. These shapes are documented with OntoInk as interactive diagrams, so that materials scientists unfamiliar with SHACL can inspect the cardinality constraints visually, experiment with instance data in the inline editor, and see the results immediately.

\begin{figure}[]
\centering
\includegraphics[width=0.7\linewidth]{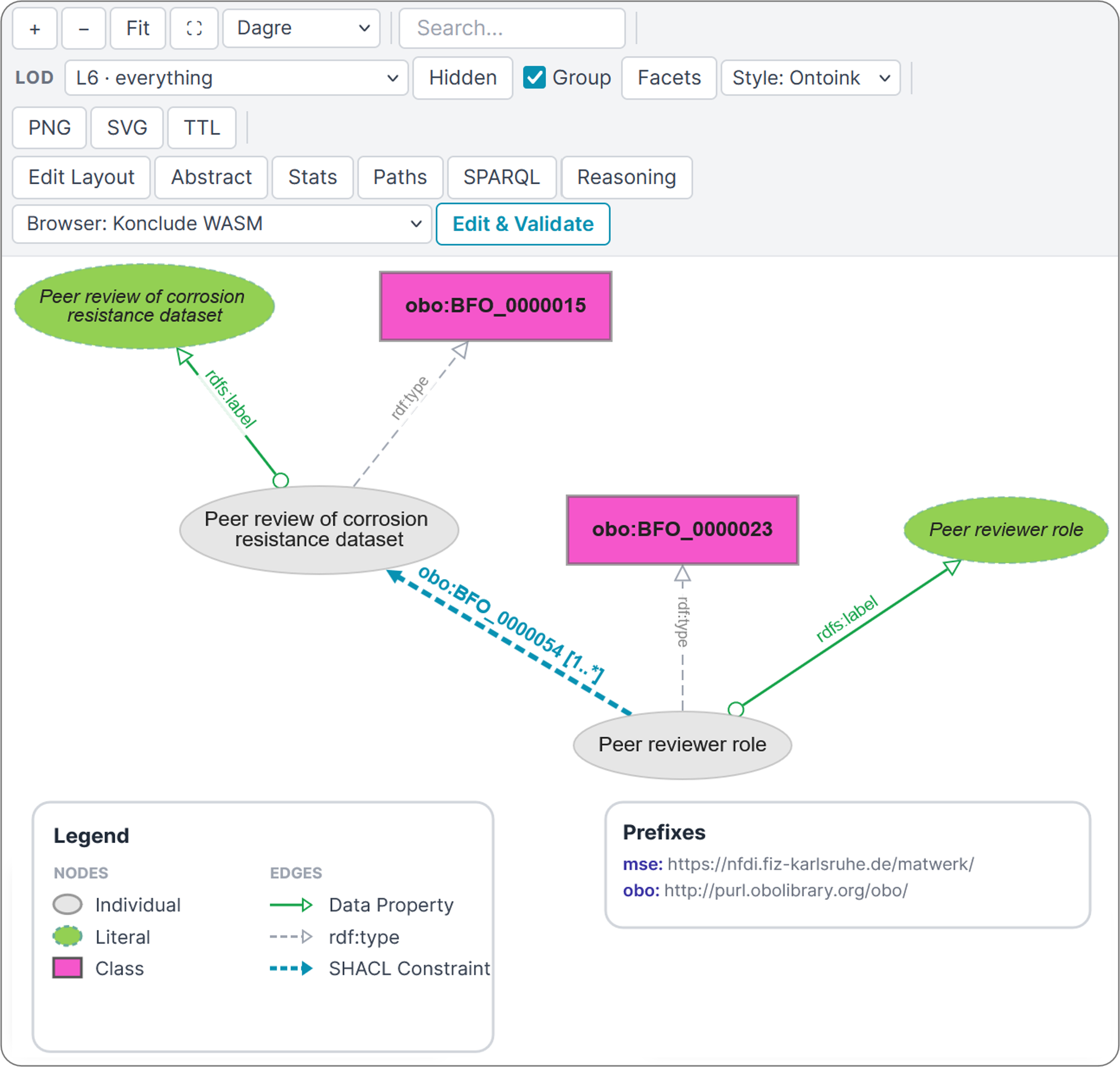}
\caption{The \emph{Role Realization} SHACL shape of the MatWerk KG, rendered by
OntoInk.}
\label{fig:matwerk-rolerealization}
\end{figure}

Figure~\ref{fig:matwerk-rolerealization} shows one of these shapes. The \emph{Role Realization} shape validates that every role must be realised in at least one process. It requires each instance of \texttt{BFO\_0000023} (\emph{role}) to carry a \texttt{BFO\_0000054} (\emph{realized in}) link to the process that brings the role into actuality, so that roles such as \emph{principal investigator} or \emph{peer reviewer} are never recorded without the process they participate in. Rendered by OntoInk, the requirement appears as a single constraint edge annotated with its cardinality: a domain expert can edit the instance data in place to watch a violation appear and disappear. The toolbar exposes what OntoInk adds over static documentation: view controls and a fuzzy search that dims non-matching elements rather than removing them. A level-of-detail (\textsf{LOD}) selector governs what is drawn, progressively admitting class hierarchy, individuals and object properties, OWL restrictions and SHACL constraints (from \textsf{L3}), and finally data properties and literals. Analysis panels report graph statistics and run reasoning over one of eight browser- or server-side backends, reporting the inferred triple count, elapsed time, and backend used, overlaying the derived triples on the graph on request, and recording every step of the run log that opens automatically when reasoning fails.

\section{Conclusion and Demonstration}\label{sec:conclusion}

OntoInk brings interactive ontology visualization, SHACL validation, OWL reasoning, and inline Turtle editing together within a documentation-as-code workflow. By treating the Turtle source as the single point of truth, it closes the gap between ontology engineering tools and documentation pipelines, whose general-purpose diagram plugins are not ontology-aware. Reasoning can be performed with HermiT~\cite{glimm2014hermit}, native Konclude~\cite{steigmiller2014konclude}, or an in-browser WebAssembly port of Konclude~\cite{hanke2026rdfreasoner} with an OWL-RL fallback, and instance data is validated against SHACL shapes both at build time with pySHACL and live in the browser.

A live demonstration is available at \url{https://ise-fizkarlsruhe.github.io/ontoink/} and includes six interactive example pages covering FOAF, Schema.org, BFO role patterns, IAO textual entities, and OWL reasoning scenarios, together with the browser-based Playground and the visual SHACL shape editor. In future work, we will (i) apply OntoInk to visualize and document developed SHACL shapes and ontology patterns; (ii) extend the Edit Layout feature with support for alternative established notations like rendering ontologies in the Graffoo~\cite{falco2014graffoo} and Chowlk~\cite{chavez2022chowlk} styles; and (iii) improve the visualization of large ontologies through clustering and graph-summarization techniques.

\begin{acknowledgments}
The authors gratefully acknowledge support from the NFDI-MatWerk consortium as part of the German National Research Data Infrastructure (NFDI) programme. NFDI is financed by the Federal Republic of Germany and the 16 federal states and funded by the Federal Ministry of Education and Research (BMBF) – funding code M532701 / the Deutsche Forschungsgemeinschaft (DFG, German Research Foundation) - project number 460247524.
\end{acknowledgments}

\begin{aideclaration}
During the preparation of this work, the author(s) used GPT-4 and Claude Sonnet~4 in order to: Grammar and spelling check. After using these tool(s)/service(s), the author(s) reviewed and edited the content as needed and take(s) full responsibility for the publication’s content.
\end{aideclaration}

\bibliography{references}

@article{alobaid2019ontoology,
  author    = {Alobaid, Ahmad and Garijo, Daniel and Poveda-Villal{\'o}n, Mar{\'\i}a and Santana-Perez, Idafen and Fern{\'a}ndez-Izquierdo, Alba and Corcho, Oscar},
  title     = {Automating Ontology Engineering Support Activities with {OnToology}},
  journal   = {Journal of Web Semantics},
  volume    = {57},
  pages     = {100472},
  year      = {2019},
  doi       = {10.1016/j.websem.2018.09.003}
}

@book{arp2015bfo,
  author    = {Arp, Robert and Smith, Barry and Spear, Andrew D.},
  title     = {{Building Ontologies with Basic Formal Ontology}},
  publisher = {MIT Press},
  address   = {Cambridge},
  year      = {2015}
}

@inproceedings{barzdins2010owlgred,
  author    = {Barzdins, Janis and Barzdins, Guntis and Cerans, Karlis and Liepins, Renars and Sprogis, Arturs},
  title     = {{OWLGrEd}: a {UML} Style Graphical Notation and Editor for {OWL} 2},
  booktitle = {Proc. of the 7th International Workshop on OWL: Experiences and Directions (OWLED 2010)},
  series    = {CEUR-WS},
  volume    = {614},
  year      = {2010}
}

@misc{beckett2014turtle,
  author    = {Beckett, David and Berners-Lee, Tim and Prud'hommeaux, Eric and Carothers, Gavin},
  title     = {{RDF} 1.1 {Turtle}},
  howpublished = {W3C Recommendation},
  year      = {2014},
  url       = {https://www.w3.org/TR/turtle/}
}

@misc{car2020pyshacl,
  author    = {Car, Nicholas J. and Homburg, Timo},
  title     = {{pySHACL}---A {Python} {SHACL} Validator},
  year      = {2020},
  url       = {https://github.com/RDFLib/pySHACL}
}

@misc{car2022pylode,
  author    = {Car, Nicholas J.},
  title     = {{pyLODE}---An {OWL} Ontology Documentation Tool},
  year      = {2022},
  url       = {https://github.com/RDFLib/pyLODE}
}

@inproceedings{chavez2022chowlk,
  author    = {Ch{\'a}vez-Feria, Serge and Garc{\'\i}a-Castro, Ra{\'u}l and Poveda-Villal{\'o}n, Mar{\'\i}a},
  title     = {Chowlk: from {UML}-Based Ontology Conceptualizations to {OWL}},
  booktitle = {The Semantic Web (ESWC 2022)},
  series    = {LNCS},
  volume    = {13261},
  pages     = {338--352},
  publisher = {Springer},
  year      = {2022},
  doi       = {10.1007/978-3-031-06981-9_20}
}

@article{dudas2018survey,
  author    = {Dud{\'a}{\v{s}}, Marek and Lohmann, Steffen and Sv{\'a}tek, Vojt{\v{e}}ch and Pavlov, Dmitry},
  title     = {{Ontology Visualization Methods and Tools: a Survey of the State of the Art}},
  journal   = {Knowledge Engineering Review},
  volume    = {33},
  pages     = {e10},
  year      = {2018},
  doi       = {10.1017/S0269888918000073}
}

@inproceedings{falco2014graffoo,
  author    = {Falco, Riccardo and Gangemi, Aldo and Peroni, Silvio and Shotton, David and Vitali, Fabio},
  title     = {Modelling {OWL} Ontologies with {Graffoo}},
  booktitle = {The Semantic Web: ESWC 2014 Satellite Events},
  series    = {LNCS},
  volume    = {8798},
  pages     = {320--325},
  publisher = {Springer},
  year      = {2014},
  doi       = {10.1007/978-3-319-11955-7_42}
}

@inproceedings{garijo2017widoco,
  author    = {Garijo, Daniel},
  title     = {{WIDOCO}: A Wizard for Documenting Ontologies},
  booktitle = {Proc. of the 16th International Semantic Web Conference (ISWC 2017)},
  series    = {LNCS},
  volume    = {10588},
  pages     = {94--102},
  publisher = {Springer},
  year      = {2017},
  doi       = {10.1007/978-3-319-68204-4_9}
}

@article{glimm2014hermit,
  author    = {Glimm, Birte and Horrocks, Ian and Motik, Boris and Stoilos, Giorgos and Wang, Zhe},
  title     = {{HermiT}: An {OWL} 2 Reasoner},
  journal   = {Journal of Automated Reasoning},
  volume    = {53},
  number    = {3},
  pages     = {245--269},
  year      = {2014}
}

@article{steigmiller2014konclude,
  author    = {Steigmiller, Andreas and Liebig, Thorsten and Glimm, Birte},
  title     = {{Konclude}: System Description},
  journal   = {Journal of Web Semantics},
  volume    = {27--28},
  pages     = {78--85},
  year      = {2014},
  doi       = {10.1016/j.websem.2014.06.003}
}

@misc{hanke2026rdfreasoner,
  author       = {Hanke, Thomas},
  title        = {{rdf-reasoner-konclude}: {OWL-DL} Tableau Reasoning in Browsers and {Node.js} via {WebAssembly}},
  year         = {2026},
  howpublished = {npm package, \url{https://www.npmjs.com/package/rdf-reasoner-konclude}},
  note         = {WASM port of Konclude~\cite{steigmiller2014konclude}}
}

@inproceedings{halilaj2016vocol,
  author    = {Halilaj, Lavdim and Petersen, Niklas and Grangel-Gonz{\'a}lez, Irlan and Lange, Christoph and Auer, S{\"o}ren and Coskun, G{\"o}khan and Lohmann, Steffen},
  title     = {{VoCol}: An Integrated Environment to Support Version-Controlled Vocabulary Development},
  booktitle = {Knowledge Engineering and Knowledge Management (EKAW 2016)},
  series    = {LNCS},
  volume    = {10024},
  pages     = {303--319},
  publisher = {Springer},
  year      = {2016},
  doi       = {10.1007/978-3-319-49004-5_20}
}

@misc{knublauch2017shacl,
  author    = {Knublauch, Holger and Kontokostas, Dimitris},
  title     = {Shapes Constraint Language ({SHACL})},
  howpublished = {W3C Recommendation},
  year      = {2017},
  url       = {https://www.w3.org/TR/shacl/}
}

@misc{krech2023rdflib,
  author    = {Krech, Daniel and others},
  title     = {{RDFLib}---A {Python} Library for Working with {RDF}},
  year      = {2023},
  url       = {https://github.com/RDFLib/rdflib}
}

@article{lamy2017owlready,
  author    = {Lamy, Jean-Baptiste},
  title     = {Owlready: Ontology-Oriented Programming in {Python} with Automatic Classification and High Level Constructs for Biomedical Ontologies},
  journal   = {Artificial Intelligence in Medicine},
  volume    = {80},
  pages     = {11--28},
  year      = {2017}
}

@article{lieber2022unshacled,
  author    = {Lieber, Sven and De Meester, Ben and Heyvaert, Pieter and Bruckmann, Florian and Wambacq, Ruben and Mannens, Erik and Verborgh, Ruben and Dimou, Anastasia},
  title     = {Visual Notations for Viewing {RDF} Constraints with {UnSHACLed}},
  journal   = {Semantic Web},
  volume    = {13},
  number    = {5},
  pages     = {757--792},
  year      = {2022},
  doi       = {10.3233/SW-210450}
}

@article{lohmann2014vowl,
  author    = {Lohmann, Steffen and Negru, Stefan and Haag, Florian and Ertl, Thomas},
  title     = {Visualizing Ontologies with {VOWL}},
  journal   = {Semantic Web},
  volume    = {7},
  number    = {4},
  pages     = {399--419},
  year      = {2016},
  doi       = {10.3233/SW-150200}
}

@inproceedings{lohmann2015webvowl,
  author    = {Lohmann, Steffen and Link, Vincent and Marber, Eduard and Negru, Stefan},
  title     = {{WebVOWL}: Web-based Visualization of Ontologies},
  booktitle = {Knowledge Engineering and Knowledge Management},
  series    = {LNCS},
  volume    = {8982},
  pages     = {154--158},
  publisher = {Springer},
  year      = {2015},
  doi       = {10.1007/978-3-319-17966-7_21}
}

@inproceedings{mouromtsev2015ontolia,
  author    = {Mouromtsev, Dmitry and Pavlov, Dmitry and Emelyanov, Yury and Morozov, Alexey and Razdyakonov, Daniil and Galkin, Mikhail},
  title     = {The Simple Web-based Tool for Visualization and Sharing of Semantic Data and Ontologies},
  booktitle = {Proc. ISWC 2015 Posters \& Demos},
  series    = {CEUR-WS},
  volume    = {1486},
  year      = {2015}
}

@article{musen2015protege,
  author    = {Musen, Mark A.},
  title     = {The {Prot{\'e}g{\'e}} Project: A Look Back and a Look Forward},
  journal   = {AI Matters},
  volume    = {1},
  number    = {4},
  pages     = {4--12},
  year      = {2015},
  doi       = {10.1145/2757001.2757003}
}

@misc{owl2profiles,
  author    = {Motik, Boris and Grau, Bernardo Cuenca and Horrocks, Ian and Wu, Zhe and Fokoue, Achille and Lutz, Carsten},
  title     = {{OWL} 2 Web Ontology Language Profiles (Second Edition)},
  howpublished = {W3C Recommendation},
  year      = {2012},
  url       = {https://www.w3.org/TR/owl2-profiles/}
}

@inproceedings{peroni2012lode,
  author    = {Peroni, Silvio and Shotton, David and Vitali, Fabio},
  title     = {The Live {OWL} Documentation Environment: A Tool for the Automatic Generation of Ontology Documentation},
  booktitle = {Proc. EKAW 2012},
  series    = {LNCS},
  volume    = {7603},
  pages     = {398--412},
  publisher = {Springer},
  year      = {2012},
  doi       = {10.1007/978-3-642-33876-2_35}
}

@inproceedings{sure2019nfdi,
  author    = {Sure-Vetter, York and Goble, Carole and Nystr{\"o}m, Mikael},
  title     = {Enabling the {National Research Data Infrastructure (NFDI)}},
  booktitle = {GI-Jahrestagung},
  pages     = {2275--2276},
  year      = {2019}
}

@inproceedings{wiens2018webvowl,
  author    = {Wiens, Vitalis and Lohmann, Steffen and Auer, S{\"o}ren},
  title     = {{WebVOWL} Editor: Device-Independent Visual Ontology Modeling},
  booktitle = {Proc. ISWC 2018 Posters \& Demonstrations},
  series    = {CEUR-WS},
  volume    = {2180},
  year      = {2018}
}

@inproceedings{duchateau2026owl2graffoo,
  author    = {Duchateau, Jakub and Debruyne, Christophe},
  title     = {{OWL2Graffoo}: Import Ontology in {diagrams.net} with {Graffoo}},
  booktitle = {The Semantic Web: ESWC 2026 Posters and Demos},
  publisher = {Springer},
  year      = {2026},
  note      = {Accepted at ESWC 2026 Posters and Demos}
}

@misc{reactodia,
  author       = {Mouromtsev, Dmitry and others},
  title        = {{Reactodia}: A {React}-Based Library for Interactive Visualization
                  and Editing of {RDF}/{OWL} Knowledge Graphs},
  year         = {2025},
  howpublished = {Open-source library, \url{https://reactodia.github.io}},
  note         = {Successor of OntoDia~\cite{mouromtsev2015ontolia}}
}

@inproceedings{gomes2024ontosurvey,
  author    = {Gomes, C. A. D. and Monfardini, G. K. d. S. Q. and Salamon, J. S. and
               Sangali, R. d. S. and Timoteo, I. S. R. and Barcellos, M. and Souza, V. E. S.},
  title     = {Investigating Tool Usage in Ontology Engineering: A Survey},
  booktitle = {Proc. 17th Seminar on Ontology Research in Brazil (ONTOBRAS 2024)},
  series    = {CEUR-WS},
  volume    = {3905},
  year      = {2024},
  url       = {https://ceur-ws.org/Vol-3905/paper6.pdf}
}

@article{mwo2026,
  author  = {Beygi Nasrabadi, Hossein and Norouzi, Ebrahim and Hubaiev, Kostiantyn
             and Waitelonis, J\"org and Sack, Harald},
  title   = {{NFDI MatWerk Ontology (MWO)}: A {BFO}-Compliant Ontology for Research
             Data Management in Materials Science and Engineering},
  journal = {Advanced Engineering Materials},
  pages   = {e202502331},
  year    = {2026},
  doi     = {10.1002/adem.202502331},
}

@misc{msekg_zenodo_v211,
  title     = {{MSE-KG} v2.1.1 -- Materials Science and Engineering Knowledge Graph},
  author    = {Norouzi, Ebrahim and Beygi Nasrabadi, Hossein and Singh, Gunjan
               and Algergawy, Alsayed and Waitelonis, J\"org and Sack, Harald},
  year      = {2026},
  publisher = {Zenodo},
  doi       = {10.5281/zenodo.19452992},
}

@software{d2,
  author       = {{Terrastruct}},
  title        = {D2: A Modern Diagram Scripting Language},
  year         = {2022},
  url          = {https://github.com/terrastruct/d2},
  note         = {GitHub repository. Accessed: 2026-07-20}
}

@software{mermaid,
  author  = {{Mermaid Contributors}},
  title   = {Mermaid: Generation of Diagrams and Flowcharts from Text in a Similar Manner as Markdown},
  year    = {2014},
  url     = {https://mermaid.js.org/},
  codeurl = {https://github.com/mermaid-js/mermaid},
  urldate = {2026-07-20}
}

\end{document}